\documentclass[letterpaper, 10 pt, journal, twoside]{IEEEtran}
\usepackage{amsmath,amsfonts}
\usepackage{algorithmic}
\usepackage{algorithm}
\usepackage{color}
\usepackage{array}
\usepackage{threeparttable}
\usepackage[caption=false,font=normalsize,labelfont=sf,textfont=sf]{subfig}
\usepackage{textcomp}
\usepackage{amssymb}
\usepackage{stfloats}
\usepackage{url}
\usepackage{verbatim}
\usepackage{graphicx}
\usepackage{cite}
\usepackage{multirow}
\usepackage{makecell}
\usepackage{booktabs}  
\usepackage{multirow}
\usepackage{threeparttable}

\begin{document}

\bstctlcite{IEEEexample:BSTcontrol}

\title{Semantic-Coupled Perception Framework for Autonomous Laparoscope Control
}

\author{Enduo Zhao, Fangzhou Li, Yujie Zhu, Sicheng Lu, Fang Chen, Longfei Ma, 

and Hongen Liao, \IEEEmembership{Senior Member,~IEEE}
\thanks{The authors acknowledge support from the National Key Researchand Development Program of China (2024YFC2418100, 2022YFC2405200), the National Natural Science Foundation of China (U22A2051), and the Natural Science Foundation of Beijing Municipality (L252201, L252126). \emph{(Correspondingauthors: Longfei Ma and Hongen Liao.)}}
\thanks{Enduo Zhao, Fangzhou Li, Yujie Zhu, Sicheng Lu, and Longfei Ma are with the School of Biomedical Engineering, Tsinghua University, Beijing, China. (e-mail: endowzhao@mail.tsinghua.edu.cn; \{lifz21, zhuyj24, lusc23\}@mails.tsinghua.edu.cn; malongfei@mail.tsinghua.edu.cn.)}%
\thanks{Fang Chen and Hongen Liao are with the School of Biomedical Engineering, Shanghai Jiao Tong University, Shanghai, China. (fchen@sjtu.edu.cn; liaohongen@sjtu.edu.cn.)}}%

\markboth{IEEE Robotics and Automation Letters.}
{Enduo Zhao \MakeLowercase{\textit{et al.}}: Semantic-Coupled Perception Framework for Autonomous Laparoscope Control} 

\maketitle

\begin{abstract}
In minimally invasive surgery (MIS), maintaining an appropriate field of view (FoV) is essential for surgical safety and efficiency. However, conventional autonomous laparoscope control methods often treat visual tracking and surgical phase recognition independently, limiting robustness under occlusion, entity overlap, and camera motion. This paper proposes a closed-loop autonomous laparoscope control framework that couples real-time surgical phase recognition with camera-motion-compensated visual perception. By exploiting phase-dependent relationships among surgical entities, the framework improves target state estimation under challenging visual conditions, supporting task-aware autonomous FoV control. Benchmark and phantom experiments evaluate the framework in surgical phase recognition, target tracking under occlusion and entity overlap, and autonomous FoV control, showing improved target visibility and task execution compared with fixed-FoV control. These results highlight the potential of incorporating task-level semantics into visual state estimation for autonomous laparoscope guidance.
\end{abstract}


\section{Introduction}

\IEEEPARstart{M}{inimally} invasive surgery (MIS) offers significant advantages over traditional open surgery, including reduced surgical trauma and faster postoperative recovery \cite{buia2015laparoscopic}. However, the limited access and indirect visualization inherent to MIS require the operative field to be visualized through a laparoscope, whose positioning is typically handled by a camera-holder assistant while the surgeon operates other instruments \cite{jarc2017viewpoint}. This arrangement introduces additional coordination demands between the surgeon and assistant and may lead to suboptimal visualization due to communication errors and involuntary camera movements \cite{gruijthuijsen2022robotic}. Ergonomic studies have further reported substantial physical burden and fatigue among camera-holder assistants during laparoscopic procedures \cite{liu2023physical}. These limitations have motivated the development of robotic autonomous laparoscope control systems that can automatically adjust the camera field of view (FoV) according to the surgical task \cite{shi2025hands}.

\begin{figure*}[!t]
\centering
\includegraphics[width=0.95\textwidth]{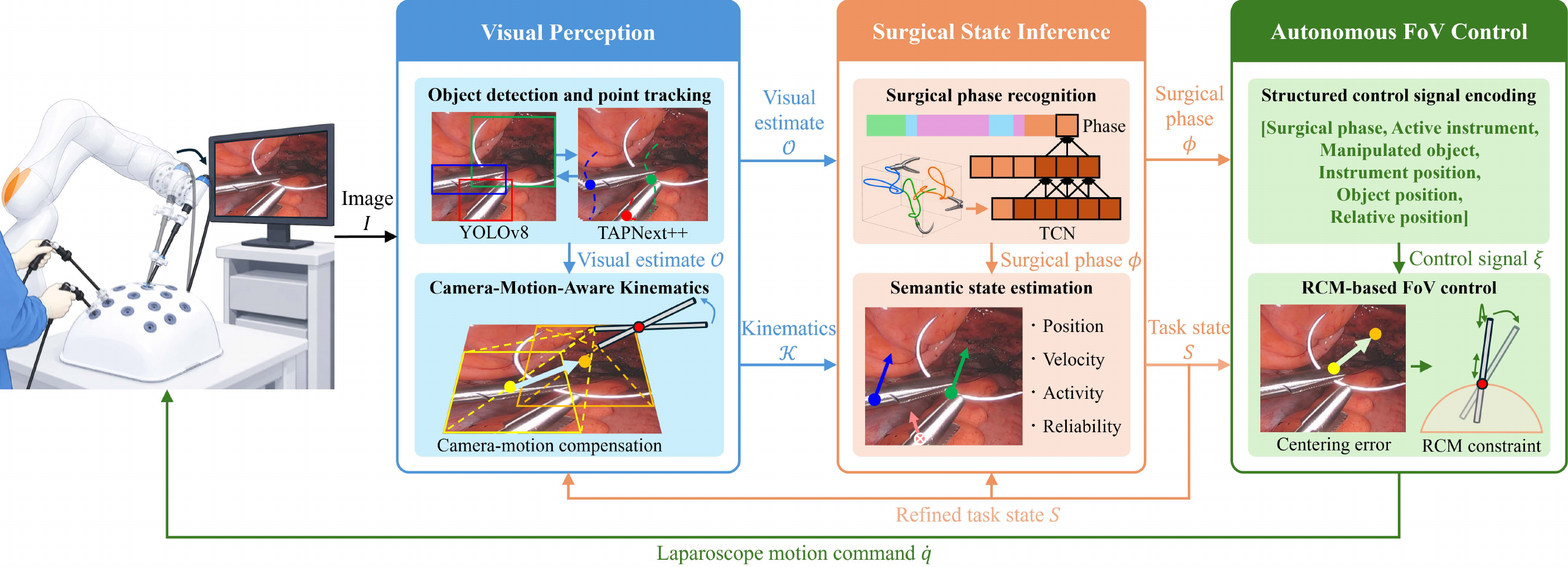}
\caption{Overview of the proposed semantic-coupled perception and control framework with three modules: Visual Perception, Surgical State Inference, and Autonomous FoV Control. Solid arrows indicate forward information transfer and feedback between the hardware system and the modules, forming a closed perception-control loop. Time-step subscripts are omitted for clarity.}
\label{fig:framework}
\end{figure*}

Current strategies for autonomous laparoscopic FoV control can be broadly categorized into direct control, instrument-based control, and task-aware approaches. Direct interfaces, such as voice commands\cite{hamza2023user}, head movements \cite{sapalidis2017laparoscopic}, or gaze-contingent control \cite{fujii2013gaze}, allow the surgeon to directly specify camera movements, but may introduce additional cognitive or operational burden by requiring explicit user commands and context switching between surgical manipulation and camera control. Instrument-based approaches instead infer camera motion from surgical tool movements observed in the laparoscopic image. Early methods relied on color markers \cite{gruijthuijsen2022robotic} or visual servoing \cite{osa2010framework}, while more recent approaches employ deep learning for instrument detection and tracking \cite{cheng2021deep,huang2022surgeon}. However, these methods often rely on predefined rules, such as following a single instrument or the midpoint between two instruments, without explicitly considering the surgical phase and the corresponding camera view requirements.

Task-aware approaches have therefore been developed to adapt laparoscope behavior according to the current surgical phase. Li et al. \cite{li2024gmm} used a Gaussian mixture model to derive phase-specific FoV requirements from instrument distributions. Sun et al. \cite{sun2025sgr} proposed SGR-AutoLap, which adjusts the FoV based on recognized surgical gestures. More recently, Zhou et al. \cite{zhou2026surglat} introduced SurgLAT, which models surgical attention as a temporally evolving latent state to infer the operative region. However, these methods primarily use task information to guide camera behavior, leaving target detection and tracking largely decoupled from the surgical phase and entity states. Beyond determining where the camera should look, the surgical phase may also constrain which entities should be tracked and how their states relate.

This opportunity arises because surgical scenes are inherently organized by the task workflow \cite{twinanda2016endonet}. During a manipulation phase, the manipulated object is spatially and semantically associated with the active instrument \cite{nwoye2022rendezvous,sharma2023surgical}, while other objects are generally uninvolved in the ongoing action. These phase-dependent relationships provide cues beyond visual appearance and short-term image motion. However, reliable target tracking remains challenging under severe occlusion \cite{chen2025research} and camera-induced image motion, which can disrupt spatial association and motion prediction. Exploiting task-dependent relational constraints together with camera motion awareness could therefore improve target state estimation and maintenance under these challenging conditions.

To address these issues, this work proposes a semantic-coupled perception framework for autonomous laparoscope control. The framework integrates object detection and tracking with real-time surgical phase recognition, where perceived states and motion relationships support phase recognition, and the recognized phase constrains object state updating. Specifically, the manipulated object is coupled with the active instrument through phase-dependent spatial relationships, while other objects are treated as non-active and semantically static. These non-active entities are used to estimate camera motion and compensate for laparoscope-induced image motion. These constraints support state maintenance, observation correction, and tracking re-initialization upon reliable detections. The resulting states are organized into a structured surgical state to determine the FoV target. Rather than replacing detection or tracking models, the framework couples their outputs with task dynamics to improve perception robustness. The framework is evaluated through ablation studies and benchmark comparisons, followed by phantom experiments on a robotic laparoscope platform using Peg Transfer and Suturing tasks to assess autonomous FoV control. The main contributions of this work are summarized as follows:

\begin{itemize}

\item A semantic-coupled perception framework integrates surgical phase recognition with object detection, point tracking, and camera-motion compensation to constrain target state estimation using phase-dependent semantics.

\item A phase-dependent state update mechanism exploits static and instrument-object binding relationships to maintain state consistency under ambiguous observations and temporary occlusions.

\item A task-aware control representation aggregates entity states according to the surgical phase for autonomous laparoscope FoV control, validated through phantom Peg Transfer and Suturing tasks.

\end{itemize}

\section{Methodology}

\subsection{Framework Overview}

As illustrated in Fig.~\ref{fig:framework}, the proposed framework establishes a semantic-coupled perception and control loop for autonomous laparoscope control. At time step \(t\), Visual Perception (Section~\ref{subsec:perception}) processes \(I_{t-1}\), \(I_t\), and the previous state \(S_{t-1}\) to obtain visual estimate \(\mathcal{O}_t\) and camera-motion-aware kinematics \(\mathcal{K}_t\). Surgical State Inference (Section~\ref{subsec:inference}) derives the phase \(\phi_t\) from the kinematic history \(\mathcal{K}_{t-\tau:t}\), obtains the corresponding relational constraints \(\mathcal{R}_t\), and integrates them with \(\mathcal{O}_t\), \(\mathcal{K}_t\), and \(S_{t-1}\) to refine the task state \(S_t\). Autonomous FoV Control (Section~\ref{subsec:FoV}) maps \(S_t\) and \(\phi_t\) to the structured control signal \(\boldsymbol{\xi}_t\) and laparoscope velocity command \(\dot{\boldsymbol{q}}_t\). The refined state is fed back as the temporal prior for the next step.

Semantic coupling arises from two temporally ordered interactions. The previous state \(S_{t-1}\) guides current-frame visual perception to improve robustness under tracking ambiguity and occlusion, whereas the phase \(\phi_t\), inferred from current kinematics, determines \(\mathcal{R}_t\) and thereby enforces task-level consistency when estimating \(S_t\). This ordering avoids a circular dependency within the same time step.

\subsection{Visual Perception}\label{subsec:perception}

Visual Perception consists of object detection and point tracking, followed by camera-motion-aware kinematics, aiming to estimate \(\mathcal{O}_t\) and \(\mathcal{K}_t\):

\begin{equation}
\left(
\mathcal{O}_t,
\mathcal{K}_t
\right)
=
\mathcal{P}
\left(
I_{t-1},I_t,S_{t-1}
\right).
\label{eq:OKPIS}
\end{equation}


\subsubsection{Object detection and point tracking}

To obtain visual estimates that are stable, spatially anchored, and temporally continuous, we combine frame-level detection with point tracking, thereby mitigating their respective limitations and enabling complementary correction.

Object detection is implemented using YOLOv8 \cite{varghese2024YOLOv8}, fine-tuned for the surgical task. The detections provide spatially anchored observations and initialize tracking points for TAPNext++ \cite{jung2026tapnext++}, which propagates these points across frames using its pretrained model without task-specific retraining. Tracked points are retained when consistent with reliable detections. When a substantial detection-tracking discrepancy occurs, detection-based correction is applied and the tracker is re-initialized for the affected entity.

Let \(\mathcal{E}\) denote the task-relevant entity set specified by the task definition, and let \(\mathcal{E}_t^{\text{est}}\subseteq\mathcal{E}\) denote the subset with valid observations at time \(t\). Let \(\mathcal{D}_t\) and \(\mathcal{T}_t\) denote the detection and tracking outputs at time \(t\), respectively. For each entity \(i\in\mathcal{E}_t^{\text{est}}\), the current visual observation is generated jointly from detection, tracking, and the previous state \(S_{t-1}\):
\begin{equation}
o_{i,t}
=
\mathcal{V}_i
\left(
\mathcal{D}_t,
\mathcal{T}_t,
S_{t-1}
\right)
=
\left(
\boldsymbol{p}_{i,t},
\text{id}_i,
\rho_{i,t}
\right),
\label{eq:oit}
\end{equation}
\noindent where \(\mathcal{V}_i(\cdot)\) denotes the observation generation function for entity \(i\), \(\boldsymbol{p}_{i,t}\in\mathbb{R}^{2}\) its current image position, \(\text{id}_i\) its identity, and $\rho_{i,t} \in [0, 1]$ the perception reliability, computed from the detection confidence and detection-tracking spatial agreement.

\(S_{t-1}\) also serves as a temporal prior for observation gating. Estimates that substantially deviate from the expected state are assigned a low \(\rho_{i,t}\) or excluded from \(\mathcal{E}_t^{\text{est}}\), thereby suppressing implausible tracking motion. The current visual observation set is defined as
\begin{equation}
\mathcal{O}_t
=
\left\{
o_{i,t}
\right\}_{i\in\mathcal{E}_t^{\text{est}}}.
\end{equation}
\begin{figure*}[!t]
\centering
\includegraphics[width=0.85\textwidth]{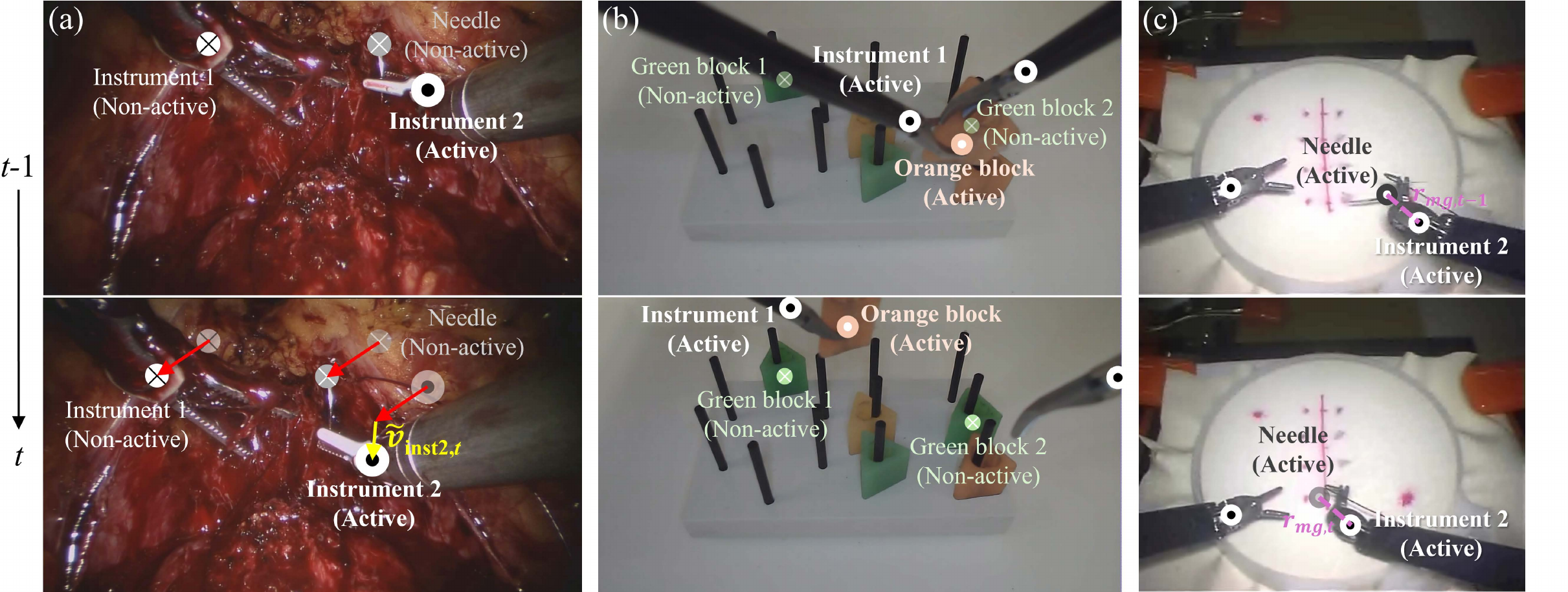}
\caption{Illustration of key components of the proposed framework.
(a) Camera-motion-aware kinematics on a suturing scene: reference features from non-active entities are used for homography estimation (red arrows), and the task-related velocity \(\tilde{\boldsymbol{v}}_{\text{inst2},t}\) (yellow arrow) of the active instrument is derived after compensation.
(b) Semantic state maintenance on a peg-transfer scene: occluded green blocks are maintained as non-active static entities and re-associated with their previous states upon reappearance.
(c) Spatial coupling on a suturing scene: the occluded needle is localized from the observable instrument 2 and the retained relative position \(\boldsymbol{r}_{mg,t-1}\), with \(\boldsymbol{r}_{mg,t}=\boldsymbol{r}_{mg,t-1}\) during occlusion.}
\label{fig:camera}
\end{figure*}


\subsubsection{Camera-motion-aware kinematics}

During autonomous laparoscope movement, stationary task entities exhibit apparent image motion, which can interfere with surgical phase recognition and semantic state estimation if directly interpreted as task-related motion; camera motion is therefore compensated before deriving task-related kinematics.

Assuming that task entities within the operative workspace lie at a locally consistent depth, inter-frame camera-induced image translation and planar motion are approximated via a 2D homography transformation. The static prior is primarily applied to rigid surgical entities, while soft-tissue deformation and respiration are beyond the current scope. To prevent task-related entity motion from corrupting the estimated transformation, the previously maintained state $S_{t-1}$ is used as a semantic prior to anchor entities that are expected to be static. Features belonging to regions associated with semantically non-active entities are preferentially selected when visible, while inconsistent or task-active regions are excluded from homography estimation. Let
\begin{equation}
\mathcal{W}_{t-1\rightarrow t}
=
\mathcal{W}
\left(
I_{t-1},
I_t,
S_{t-1}
\right)
\end{equation}
denote the estimated image transformation caused by laparoscope motion between two consecutive frames. The previous image position \(\hat{\boldsymbol{p}}_{i,t-1}\) of entity \(i\) is obtained from \(S_{t-1}\) (detailed in Eq.~\eqref{eq:sit}) and transformed into the current image coordinate system as
\begin{equation}
\hat{\boldsymbol{p}}_{i,t}^{c}
=
\mathcal{W}_{t-1\rightarrow t}
\left(
\hat{\boldsymbol{p}}_{i,t-1}
\right),
\label{eq:pitc}
\end{equation}
where \(\hat{\boldsymbol{p}}_{i,t}^{c}\) denotes the camera-motion-compensated prediction of the entity position. The task-related image velocity is then calculated as
\begin{equation}
\tilde{\boldsymbol{v}}_{i,t}
=
\frac{
\boldsymbol{p}_{i,t}
-
\hat{\boldsymbol{p}}_{i,t}^{c}
}{
\Delta t
}.
\label{eq:hatvit}
\end{equation}

As a representative example, Fig.~\ref{fig:camera}(a) illustrates this process on a suturing scene from the publicly available SAR-RARP50 dataset \cite{psychogyios2023sar}.

In addition to individual entity motion, spatial relationships between relevant entities are extracted. For entities \(i\) and \(j\), their observation-level relative image position is defined as
\begin{equation}
\boldsymbol{r}^{\text{obs}}_{ij,t}
=
\boldsymbol{p}_{i,t}
-
\boldsymbol{p}_{j,t}.
\end{equation}

The camera-motion-aware kinematics is consequently defined as
\begin{equation}
\mathcal{K}_t
=
\left\{
\tilde{\boldsymbol{v}}_{i,t}
\right\}_{i\in\mathcal{E}_t^{\text{est}}}
\;\cup\;
\left\{
\boldsymbol{r}^{\text{obs}}_{ij,t}
\right\}_{(i,j)\in\mathcal{E}_t^{\text{pair}}},
\end{equation}

\noindent where \(\mathcal{E}_t^{\text{pair}} \subseteq \{(i,j)\mid i,j\in\mathcal{E}_t^{\text{est}},\,i\neq j\}\) denotes the task-relevant entity pairs considered at time \(t\). Which entities are paired and how each pair is ordered are specified by the task definition; each ordered pair \((i,j)\in\mathcal{E}_t^{\text{pair}}\) therefore represents a directed, semantically meaningful spatial relation.

The two outputs of Visual Perception therefore serve different purposes. \(\mathcal{O}_t\) provides the visual estimate for state anchoring and perception reliability assessment, while \(\mathcal{K}_t\) provides the temporal and relational information required for surgical phase recognition and semantic consistency evaluation.

\subsection{Surgical State Inference}\label{subsec:inference}

Surgical State Inference comprises surgical phase recognition and semantic state estimation, aiming to obtain the current surgical phase $\phi_t$ and the refined state $S_t$.


\subsubsection{Surgical phase recognition}

The current surgical phase is estimated from a temporal window of kinematics:

\begin{equation}
\phi_t = \mathcal{H}(\mathcal{K}_{t-\tau:t}),
\end{equation}

\noindent where \(\tau\) denotes the window length and \(\mathcal{H}\) is a task-specific temporal convolutional network (TCN) \cite{bai2018empirical}  trained on the corresponding kinematic sequences after camera-motion compensation. The variable entity set \(\mathcal{K}_t\) is mapped to a fixed-order feature vector, with unobserved entities zero-padded and a binary presence mask.

\subsubsection{Semantic state estimation}

The semantic state estimation step estimates the refined state by
\begin{equation}
S_t
=
\mathcal{U}
\left(
S_{t-1},
\mathcal{O}_t,
\mathcal{K}_t,
\mathcal{R}_t
\right),
\label{eq:St}
\end{equation}
where \(\mathcal{R}_t\) denotes the phase-dependent relational constraints used to maintain task-level consistency.

To bootstrap this recursive process, the initial state $S_0$ is directly constructed from the YOLOv8 detections and TAPNext++ tracking points in the first frame, with the surgical phase initialized to the default starting phase $\phi_0$ (e.g., Idle).

\paragraph{Phase-dependent relational constraints}\label{para:rel}

The semantic relationships activated by the current surgical phase are represented as
\begin{equation}
\mathcal{R}_t
=
\Psi_{\text{sem}}
\left(
\phi_t,\mathcal{E}
\right),
\end{equation}
where \(\mathcal{E}\) is the task-relevant entity set. The function \(\Psi_{\text{sem}}\) provides a task-independent representation mechanism, while the phase-to-relationship mapping is instantiated for each task. 


The output \(\mathcal{R}_t\) encodes the phase-activated relational structure. For each active manipulation, a relation \(\mathtt{manipulates}(\mathcal{C}_{m},\mathcal{C}_{g})\) identifies the candidate sets \(\mathcal{C}_{m},\mathcal{C}_{g}\subseteq\mathcal{E}\) for the active instrument and the manipulated object, and specifies the constraint between them. Hereafter, \(m_t\in\mathcal{C}_m\) denotes the active instrument specified by the current phase, and \(g_t\in\mathcal{C}_g\) the manipulated object that remains relatively static with respect to it.

Two representative relational primitives are considered. First, an entity not involved in the current manipulation is treated as semantically non-active, with its task-related velocity expected to remain close to zero (\(\tilde{\boldsymbol{v}}_{i,t}\approx\mathbf{0}\)). Second, when an instrument actively manipulates an object, their states are coupled by a relative-position constraint. These two primitives provide complementary motion and spatial consistency cues for state estimation. For example, Fig.~\ref{fig:camera}(b) shows a peg-transfer scene from the Peg Transfer dataset \cite{Lau2025}, where the block manipulated by the active instrument is represented by a coupled relationship, while the remaining blocks are treated as non-active entities. 


\paragraph{State update}

The maintained task state \(S_t\) consists of individual entity states and inter-entity relative positions:
\begin{equation}
S_t =
\left(
\left\{\boldsymbol{s}_{i,t}\right\}_{i\in\mathcal{E}},
\left\{\boldsymbol{r}_{ij,t}\right\}_{(i,j)\in\mathcal{E}^{\text{pair}}}
\right).
\end{equation}
Each entity state is defined as
\begin{equation}
\boldsymbol{s}_{i,t}
=
\left(
\hat{\boldsymbol{p}}_{i,t},
\tilde{\boldsymbol{v}}_{i,t},
\alpha_{i,t},
\rho_{i,t}
\right),
\label{eq:sit}
\end{equation}
where \(\hat{\boldsymbol{p}}_{i,t}\) is the maintained image position after semantic fusion (detailed below), \(\tilde{\boldsymbol{v}}_{i,t}\) the task-related image velocity (see Eq.~\eqref{eq:hatvit}), \(\rho_{i,t}\) the perception reliability (see Eq.~\eqref{eq:oit}), and \(\alpha_{i,t}\in\{0,1\}\) the task activation state specified by \(\mathcal{R}_t\), with \(1\) and \(0\) indicating active and non-active entities, respectively.

The relative position for relevant entity pairs is estimated as
\begin{equation}
\boldsymbol{r}_{ij,t}
=
\begin{cases}
\hat{\boldsymbol{p}}_{i,t}-\hat{\boldsymbol{p}}_{j,t},
& (i,j)\in\mathcal{E}^{\text{pair}}_t,\\[2pt]
\boldsymbol{r}_{ij,t-1},
& \text{otherwise}.
\end{cases}
\label{eq:rij_update}
\end{equation}
When no semantic correction is applied to either entity, the maintained relative position coincides with the observation-level estimate, i.e., \(\boldsymbol{r}_{ij,t}=\boldsymbol{r}^{\text{obs}}_{ij,t}\). Otherwise, the previous relative position is retained for occlusion handling.








\paragraph{State maintenance and recovery}

When an entity becomes temporarily occluded or its visual observation becomes unreliable, its state is maintained according to its semantic role and available relational constraints.

If the occluded entity is non-active and expected to remain stationary, its position is maintained using the camera-motion-compensated prediction (defined by Eq.~\eqref{eq:pitc}), with
\begin{equation}
\hat{\boldsymbol{p}}_{i,t}
=
\hat{\boldsymbol{p}}_{i,t}^{c},
\qquad
\tilde{\boldsymbol{v}}_{i,t}
=
\mathbf{0}.
\end{equation}

If the occluded entity is active and has an established coupling \((m_t,g_t)\) with an observable counterpart, its position is inferred from the observable entity and the relative position retained from \(S_{t-1}\):
\begin{equation}
\hat{\boldsymbol{p}}_{m,t}-\hat{\boldsymbol{p}}_{g,t}=\boldsymbol{r}_{mg,t-1}.
\end{equation}
If its coupled counterpart is also unobservable, or if the active entity has no established coupling (e.g., an instrument without a manipulated object), its state is temporarily maintained using the TAPNext++ tracking output.

Once reliable visual observations are restored, the maintained states are re-anchored to the new observations, and the tracker is re-initialized as needed. After an extended FoV absence, a reappearing target is matched to the original entity (rather than treated as new) only if the detection falls within a search radius around the camera-compensated prediction \(\hat{\boldsymbol{p}}_{i,t}^{\text{c}}\) and its appearance similarity to the historical state exceeds a threshold; otherwise, the prior state is discarded. Moreover, TAPNext++ propagation is suspended if an entity remains unobserved and excluded from \(\mathcal{E}_t^{\text{est}}\) beyond a temporal limit. Via the temporal feedback in Eq.~\eqref{eq:OKPIS}, \(S_t\) serves as the prior to support state maintenance and recovery in the next perception step.

Fig.~\ref{fig:camera}(b) illustrates the maintenance of non-active blocks, while Fig.~\ref{fig:camera}(c) shows a suturing scene from the JIGSAWS dataset \cite{gao2014jhu} to illustrate the semantic coupling between an active instrument and a manipulated object during occlusion.

\subsection{Autonomous FoV Control}\label{subsec:FoV}

Autonomous FoV Control, consisting of  structured control signal encoding and remote center of motion (RCM)-based FoV control, outputs the laparoscope motion command \(\dot{\boldsymbol{q}}_t\).

\subsubsection{Structured control signal encoding}
\label{subsec:encoding}

The refined task state \(S_t\) and current phase \(\phi_t\) are mapped to a compact, control-oriented representation by selecting only the entities that determine the laparoscope FoV:

\begin{equation}
\boldsymbol{\xi}_t
=
\mathcal{G}
\left(
S_t,\phi_t
\right)
=
\left(
\phi_t,
\mathcal{A}_t,
\left\{
\hat{\boldsymbol{p}}_{i,t}
\right\}_{i\in\mathcal{A}_t}
\right),
\end{equation}
where \(\mathcal{A}_t\) denotes the set of active entities participating in the current manipulation, and \(\hat{\boldsymbol{p}}_{i,t}\) their refined image positions. This representation avoids passing the full maintained state to the controller and focuses it on the current manipulation context. The set \(\mathcal{A}_t\) is determined by
\begin{equation}
\mathcal{A}_t
=
\left\{
i \in \mathcal{E} : \alpha_{i,t} = 1
\right\},
\end{equation}
where \(\alpha_{i,t}\) is the task activation state. Since a single phase may involve multiple instruments and manipulated objects, \(\mathcal{A}_t\) includes all active entities rather than a single target.

\begin{table*}[t]
\centering
\small 
\setlength{\tabcolsep}{2.5pt} 

\caption{Ablation study of target tracking and surgical phase recognition.}
\label{tab:tracking_ablation}
\begin{tabular}{ccccccccccccccccccc}
\toprule
\multicolumn{4}{c}{Components} & \multicolumn{5}{c}{SAR-RARP50 \cite{psychogyios2023sar}} & \multicolumn{5}{c}{Peg Transfer \cite{Lau2025}} & \multicolumn{5}{c}{JIGSAWS \cite{gao2014jhu}} \\
\cmidrule(lr){1-4} \cmidrule(lr){5-9} \cmidrule(lr){10-14} \cmidrule(lr){15-19}

\multirow{2}{*}{Detection} & \multirow{2}{*}{Tracking} & \multirow{2}{*}{\makecell{Camera\\Comp.}} & \multirow{2}{*}{\makecell{Semantic\\Coupling}} & 
\multicolumn{3}{c}{Tracking} & \multicolumn{2}{c}{Phase} & 
\multicolumn{3}{c}{Tracking} & \multicolumn{2}{c}{Phase} & 
\multicolumn{3}{c}{Tracking} & \multicolumn{2}{c}{Phase} \\
\cmidrule(lr){5-7} \cmidrule(lr){8-9} \cmidrule(lr){10-12} \cmidrule(lr){13-14} \cmidrule(lr){15-17} \cmidrule(lr){18-19}

& & & & 
MPE & IDS & ORL & AC & JA & 
MPE & IDS & ORL & AC & JA & 
MPE & IDS & ORL & AC & JA \\
\midrule
\checkmark & --         & --         & --         & 21.7 & 23.1 & 13.0 & 62.7 & 47.2 & 52.2 & 41.4  & 27.6  & 71.4 & 61.0 & 35.5 & 26.1  & 14.1  & 67.5 & 50.1 \\
\checkmark & \checkmark & --         & --         & 13.2 & 11.4 & 7.6  & 66.9 & 54.3 & 36.9 & 19.5  & 12.3  & 79.5 & 67.6 & 21.3 & 20.5  & 10.3  & 72.3  & 59.4 \\
\checkmark & \checkmark & \checkmark & --         & 9.4  & 9.9  & 6.7  & 82.3 & 74.1 & N/A  & N/A & N/A & N/A  & N/A  & N/A  & N/A & N/A & N/A  & N/A  \\
\checkmark & \checkmark & \checkmark & \checkmark & \textbf{2.5}  & \textbf{4.0}  & \textbf{5.2}  & \textbf{89.9} & \textbf{79.0} & \textbf{7.1}  & \textbf{5.8}   & \textbf{5.5}   & \textbf{91.7} & \textbf{83.8} & \textbf{4.2}  & \textbf{7.0}   & \textbf{6.8}   & \textbf{90.1} & \textbf{81.2} \\
\bottomrule
\end{tabular}
\end{table*}

\subsubsection{RCM-based FoV control}





Using \(\boldsymbol{\xi}_t\) as input, a representative target position \(\boldsymbol{p}^{*}_{t}\) is computed by phase-dependent aggregation of the refined positions of all active entities in \(\mathcal{A}_t\):

\begin{equation}
\boldsymbol{p}^{*}_{t}
=
\mathcal{F}
\left(
\left\{
\hat{\boldsymbol{p}}_{i,t}
\right\}_{i\in\mathcal{A}_t},
\phi_t
\right),
\label{eq:ptxing}
\end{equation}
where \(\mathcal{F}\) denotes a general phase-dependent aggregation operator (e.g., weighted centroid) customizable to task-specific viewing preferences. 

The image-space centering error relative to the image center \(\boldsymbol{p}_{c}\) is defined as \(\boldsymbol{e}_t=\boldsymbol{p}^{*}_{t}-\boldsymbol{p}_{c}\). The laparoscope motion command is then generated from the centering error while satisfying the kinematic constraints of the robotic platform:

\begin{equation}
\dot{\boldsymbol{q}}_t
=
\mathcal{C}_{\text{RCM}}
\left(
\boldsymbol{e}_t,
\boldsymbol{q}_t
\right),
\quad
\text{s.t.}
\quad
\boldsymbol{J}_{\text{RCM}}
\left(
\boldsymbol{q}_t
\right)
\dot{\boldsymbol{q}}_t
=
\boldsymbol{0}.
\label{eq:qdot}
\end{equation}

In practice, \(\mathcal{C}_{\text{RCM}}\) is implemented as a QP tracking the proportional law \(\boldsymbol{v}_t=-k_p\boldsymbol{e}_t\) by minimizing \(\|\boldsymbol{J}_{\text{img}}(\boldsymbol{q}_t)\dot{\boldsymbol{q}}-\boldsymbol{v}_t\|_2^2+\lambda\|\dot{\boldsymbol{q}}\|_2^2\) with \(\lambda>0\) subject to the RCM constraint and the joint position and velocity limits, solved using OpTaS and CasADi. The executed motion generates the next camera observation \(I_{t+1}\), thereby closing the perception-control loop.

\section{Experiments}

The proposed framework is evaluated through an ablation study on target tracking and surgical phase recognition (Experiment~I), a comparison with state-of-the-art (SOTA) phase recognition methods (Experiment~II), and phantom validation of autonomous FoV control (Experiment~III).

\subsection{Parameter Settings}

Except for the settings inherited from the pretrained detection and tracking models, the thresholds below were manually tuned on the validation splits. All experiments process \(1280\times720\) px videos at \(60\) Hz. Observation gating rejects estimates deviating from the prior position by more than \(80\) px or with reliability below \(0.40\); entities are treated as static when the compensated velocity is below \(15\) px/s; instrument--object couplings are established when the inter-entity distance is below \(45\) px; and the phase-recognition window is \(\tau=90\) frames ($3.0$ s). Re-identification uses a search radius of \(60\) px and an appearance similarity of \(0.75\), and tracking is suspended after \(5\) unobserved frames. For FoV control, the proportional gain is \(k_p=0.8\), the regularization coefficient is \(\lambda=0.05\), and the joint velocity limit is \(0.25\) rad/s.

\subsection{Experiment I: Ablation Study on Target Tracking and Surgical Phase Recognition}

\subsubsection{Configurations}

To evaluate the contribution of each key component of the proposed framework, four configurations are compared incrementally: 1) detection only (YOLOv8); 2) detection + tracking (TAPNext++); 3) detection + tracking + camera compensation; 4) full method.

\subsubsection{Datasets and metrics}

Three datasets are selected to evaluate complementary capabilities of the proposed perception module. SAR-RARP50 \cite{psychogyios2023sar} features substantial laparoscope viewpoint changes, testing camera-motion compensation. Peg Transfer \cite{Lau2025} involves frequent block overlap, challenging entity discrimination under partial occlusion. The suturing subset of JIGSAWS \cite{gao2014jhu} features needles frequently leaving and re-entering the FoV, testing semantic state maintenance and re-anchoring. As Peg Transfer and JIGSAWS contain negligible laparoscope motion, camera-motion compensation is evaluated only on SAR-RARP50.

The ablation is evaluated on two complementary aspects. For target tracking, we report image-space mean position error (MPE, pixels), identity switches (IDS, counts), and occlusion recovery latency (ORL, frames), evaluated against groundtruth manually annotated and cross-checked by two annotators. For surgical phase recognition, we report video-level accuracy (AC, \%) and phase-level Jaccard coefficient (JA, \%) using the ground-truth phase labels provided by the datasets.

\subsubsection{Results}

Table~\ref{tab:tracking_ablation} summarizes the ablation results. Adding TAPNext++ consistently reduces MPE, IDS, and ORL compared with detection, demonstrating improved temporal continuity. Camera-motion compensation further improves both tracking and phase recognition on SAR-RARP50. The full method achieves the best overall performance on all three datasets, with particularly strong results on the Peg Transfer and JIGSAWS datasets under occlusion and overlap. These results indicate that temporal tracking, camera-motion compensation, and phase-dependent semantic constraints play complementary roles in robust target state estimation.


\begin{table}[t]
\centering
\footnotesize
\setlength{\tabcolsep}{3pt}
\caption{Comparison with state-of-the-art phase recognition methods}
\label{tab:phase_comparison}
\begin{tabular}{llcccc}
\toprule
Dataset & Method & AC (\%) & PR (\%) & RE (\%) & JA (\%) \\
\midrule
\multirow{5}{*}{SAR-RARP50 \cite{psychogyios2023sar}} 
 & NFSM \cite{ding2024neural}     &  82.5   &  80.1   &  79.5   &  71.1   \\
 & MS-ASCT \cite{zhang2024friends} &  86.9   &  84.1 &  84.0   &  75.5   \\
 & STSANet \cite{li2025stsanet} &  88.4   &  86.7   &  86.6   &  77.2   \\
 & LemonFM \cite{che2026lemon}     &  89.1   &  88.3  &  \textbf{88.1}   &  78.7   \\
 & Ours                            & \textbf{89.9} & \textbf{88.3} & 88.0 & \textbf{79.0} \\
\midrule
\multirow{5}{*}{Peg Transfer \cite{Lau2025}} 
 & NFSM \cite{ding2024neural}     &  78.2   &  77.8   &  77.7   &  61.7   \\
 & MS-ASCT \cite{zhang2024friends} &  80.4   &  79.1   &  79.1   &  66.2   \\
 & STSANet \cite{li2025stsanet} &  81.1   &  80.1   &  79.8   &  67.3   \\
 & LemonFM \cite{che2026lemon}     &  83.8   & 82.8   & 82.4   &  70.8   \\
 & Ours                            & \textbf{91.7} & \textbf{91.6} & \textbf{90.7} & \textbf{83.7} \\
\midrule
\multirow{5}{*}{JIGSAWS \cite{gao2014jhu}} 
 & NFSM \cite{ding2024neural}     &  86.6   &  81.7   &  81.0   &  70.5   \\
 & MS-ASCT \cite{zhang2024friends} &  87.1   &  83.4   &  82.9   &  73.3   \\
 & STSANet \cite{li2025stsanet} &  89.0   &  87.6   &  87.4   &  79.5   \\
 & LemonFM \cite{che2026lemon}     &  \textbf{90.7}   &  88.1   &  87.9   &  \textbf{81.8}   \\
 & Ours                            & 90.1 & \textbf{88.7} & \textbf{88.4} & 81.2 \\
\bottomrule
\end{tabular}
\end{table}

\begin{figure}[!t]
\centering
\includegraphics[width=0.4\textwidth]{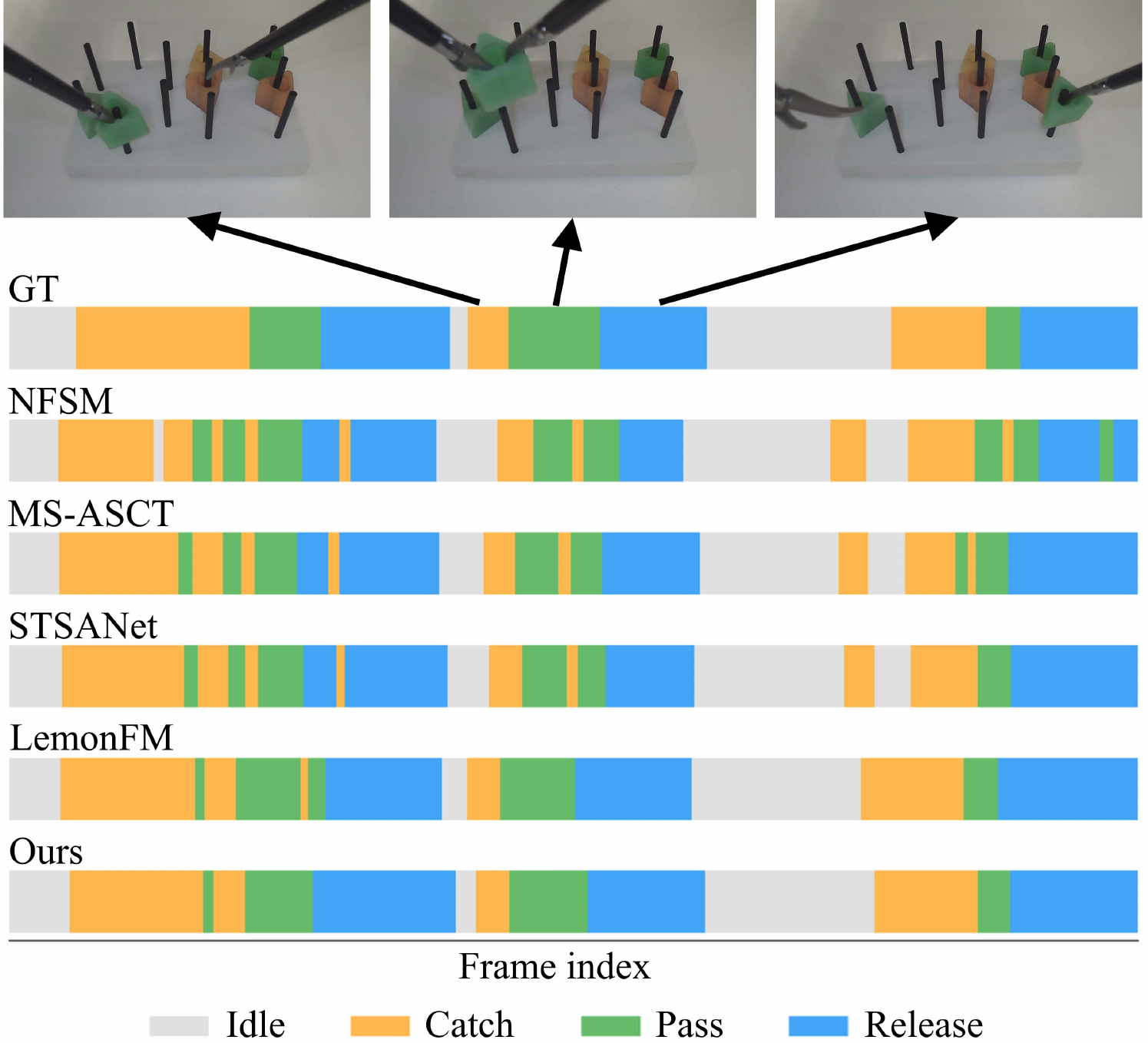}
\caption{Qualitative comparison with SOTA methods on the Peg Transfer dataset. Different colors represent different surgical phases.}
\label{fig:phase_qualitative}
\end{figure}

\subsection{Experiment II: Comparison with SOTA Surgical Phase Recognition Methods}

\subsubsection{Setups}

To evaluate the effectiveness of the proposed Surgical State Inference for surgical phase recognition, four recent SOTA methods are compared: NFSM \cite{ding2024neural}, MS-ASCT \cite{zhang2024friends}, STSANet \cite{li2025stsanet}, and LemonFM \cite{che2026lemon}. 

Besides the metrics used in Experiment~I, phase-level precision (PR) and recall (RE) are also reported to evaluate the accuracy and completeness of surgical phase recognition.

The same three datasets in Experiment I are used. All methods are trained and evaluated using the same data split for each dataset to ensure a fair comparison.

\subsubsection{Results}

Table~\ref{tab:phase_comparison} summarizes the comparison results. On SAR-RARP50, the proposed method achieves the highest AC and JA, with PR matching the best baseline and RE slightly lower; this gain is mainly attributed to camera-motion-aware kinematics under substantial viewpoint changes. On the Peg Transfer dataset, it outperforms all baselines on all four metrics, with a clear margin under frequent entity overlap. Fig.~\ref{fig:phase_qualitative} further presents a qualitative comparison of phase recognition on Peg Transfer, where our method yields phase transitions closer to the ground truth, with fewer fragmented predictions and spurious transitions around entity overlap. This improved temporal consistency is mainly due to the phase-dependent relational constraints. On JIGSAWS, the method achieves the highest PR and RE while remaining competitive in AC and JA, with only a marginal gap to LemonFM.

\begin{figure}[!t]
\centering
\includegraphics[width=0.48\textwidth]{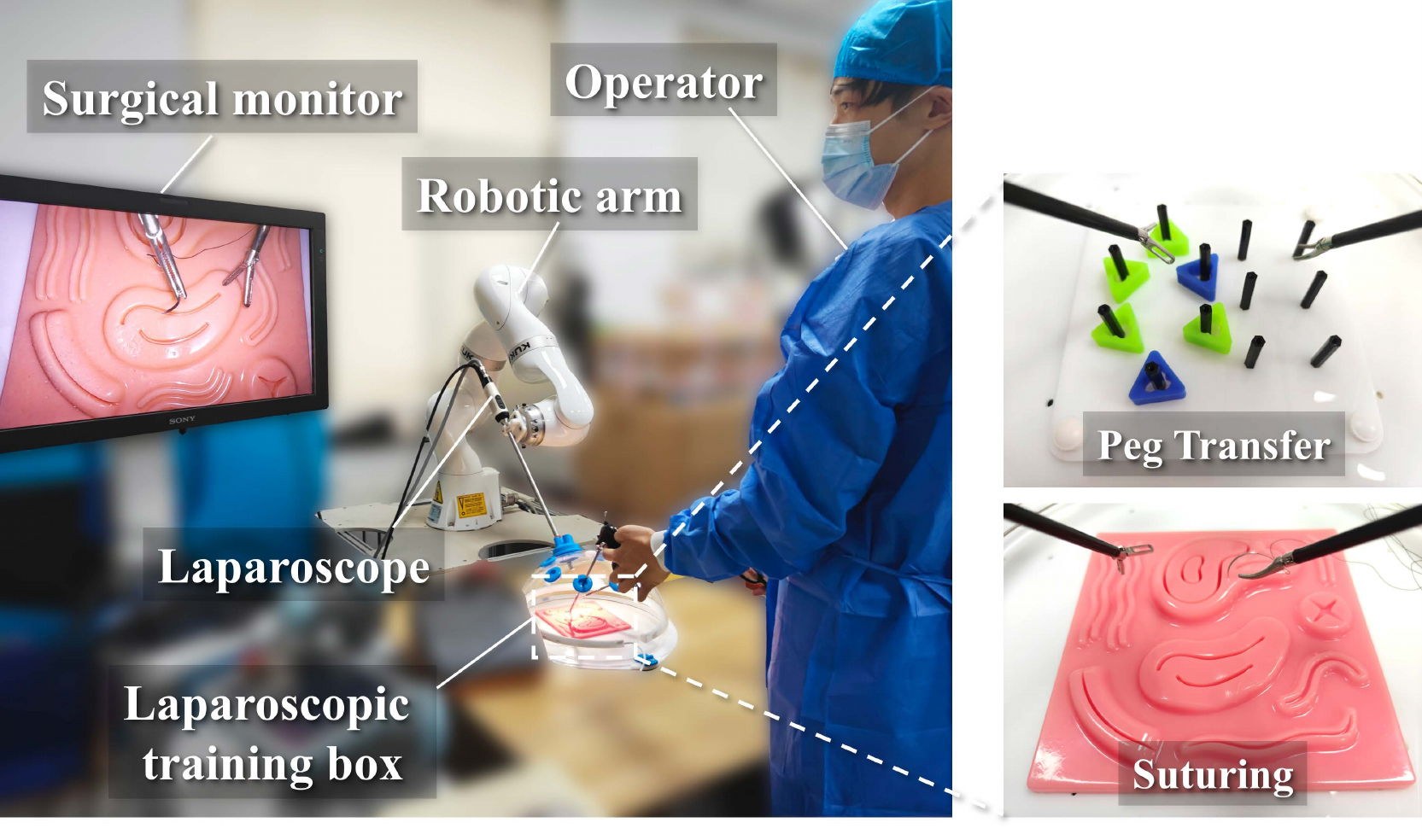}
\caption{Hardware setup of Experiment III consists of a robotic arm holding
a laparoscope, a surgical monitor, and a laparoscopic training box with standard FLS Peg Transfer and Suturing modules.}
\label{fig:hardware}
\end{figure}

\begin{figure*}[t]
    \centering
    \includegraphics[width=\textwidth]{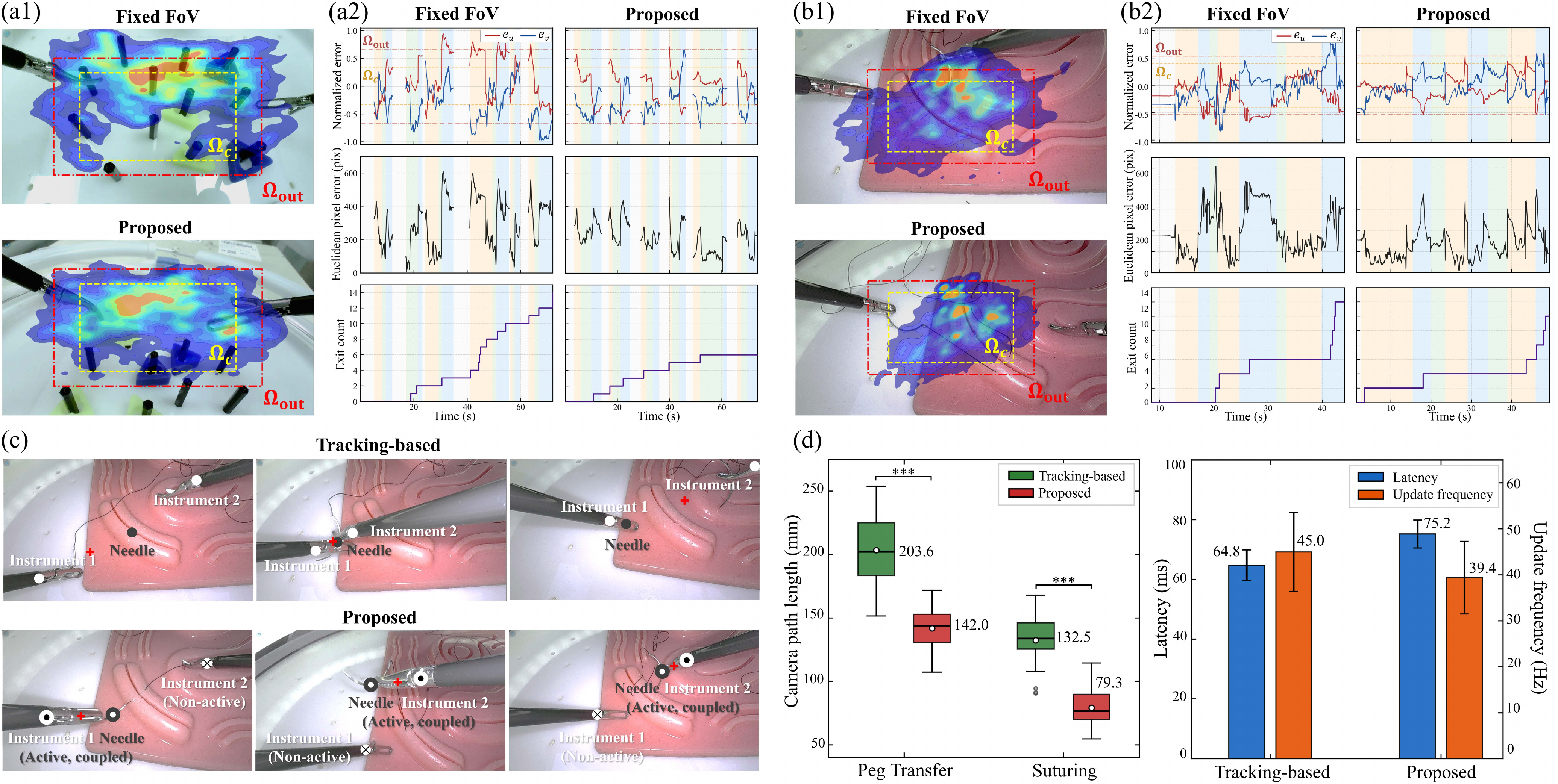}
    \caption{Results of Experiment~III. (a1)(b1) Spatial heatmaps of active entity positions in Peg Transfer and Suturing under fixed FoV and the proposed control, with the safe region \(\Omega_c\) (yellow box) and outer triggering region \(\Omega_{\text{out}}\) (red box). (a2)(b2) Temporal profiles of horizontal/vertical errors (\(e_u\), \(e_v\), normalized), Euclidean pixel error, and cumulative exit count over representative trials. (c) Tracking-based versus proposed control in a needle-passing scene during Suturing, showing the FoV target (red cross) and entity states. (d) Camera path length, control latency, and update frequency in both tasks.}
    \label{fig:exp3_results}
\end{figure*}

\subsection{Experiment III: Phantom Validation on Peg Transfer and Suturing}
\label{sec:exp3}

\subsubsection{Experimental protocol}
\label{subsec:exp3_protocol}

This experiment evaluates the framework in two standard Fundamentals of Laparoscopic Surgery (FLS) phantom tasks: Peg Transfer (transferring six blocks between opposite pegs) and Suturing (performing three interrupted sutures). Three conditions are compared: fixed-FoV control, tracking-based FoV control, and the proposed semantic-coupled control. Here, the tracking-based condition serves as a decoupled task-aware baseline: it retains the same phase recognizer, phase-selected target aggregation rule, camera-motion compensation, and RCM controller as the proposed method, but directly uses detector and tracker outputs to compute the FoV target, bypassing our perception-level relational state update and occlusion-handling mechanisms. Four operators, including novices and experts, each perform five trials per task under each condition, yielding \(N=20\) trials per condition for each task. The condition order is counterbalanced across operators.

\subsubsection{Hardware setup}
\label{subsec:exp3_hardware}

As illustrated in Fig.~\ref{fig:hardware}, the experimental platform consists of a 7-DoF robotic arm (LBR iiwa 7 R800 CR, KUKA, Germany), a laparoscope (3DWFEL30-G2, MedBot, China), a surgical monitor (LMD-3251TC, SONY, Japan), and a laparoscopic training box with standard FLS Peg Transfer and Suturing modules. The laparoscope is calibrated (intrinsics and hand-eye) before the experiments. All perception, recognition, and control modules run on a laptop equipped with a GeForce RTX 5070 Ti Laptop GPU (NVIDIA, USA).

\subsubsection{Proposed FoV control strategy}
\label{subsec:exp3_strategy}

The proposed semantic-coupled autonomous FoV control follows the strategy described in Section~\ref{subsec:FoV}. While \(\mathcal{F}\) in Eq.~\eqref{eq:ptxing} generally accommodates phase-specific weighting strategies, in our experiments it is instantiated as the centroid of the active entity set \(\mathcal{A}_t\), allowing the FoV target to adapt across surgical phases. The controller \(\mathcal{C}_{\text{RCM}}\) is event-triggered: the laparoscope remains stationary while the target stays within a central safe region \(\Omega_c\), and starts moving when the target exits an outer triggering region \(\Omega_{\text{out}}\), where \(\Omega_c \subset \Omega_{\text{out}}\). The camera stops once the target returns to \(\Omega_c\). While the target lies between the two regions, the previous trigger state is retained to avoid chattering. Both regions are rectangular, covering the central \(1/2\), \(2/3\) of the image dimensions for Peg Transfer, and \(2/5\), \(8/15\) for Suturing. To ensure operational safety, \(\mathcal{C}_{\text{RCM}}\) enforces the joint position and velocity limits in the QP, holds the camera stationary if all active entities lose reliable state estimates, and allows immediate operator override.

\subsubsection{Metrics}
\label{subsec:exp3_metrics}

The evaluation considers four aspects:

\begin{itemize}
    \item \textbf{Visual tracking and FoV control:} horizontal and vertical image-space errors \(e_u\) and \(e_v\) (pixels), Euclidean pixel error \(e_{\text{pix}}=\sqrt{e_u^2+e_v^2}\) (pixels), proportion of time the target remains within \(\Omega_c\) (\%), and the number of exits from \(\Omega_{\text{out}}\) (counts).
    
    \item \textbf{Task performance:} task completion time (s).
    
    \item \textbf{Camera motion:} cumulative camera path length (mm).
    
    \item \textbf{System performance:} latency (ms) and end-to-end update frequency (Hz) of FoV control commands.
\end{itemize}

\begin{table}[t]
    \centering
    \footnotesize
    \setlength{\tabcolsep}{3.5pt} 
    \caption{Quantitative results of fixed, tracking-based, and semantic-coupled FoV control.}
    \label{tab:exp3_metrics}
    \begin{tabular}{llccc}
        \toprule
        Task & Metric & Fixed & Tracking & Proposed \\
        \midrule
        \multirow{4}{*}{\makecell[l]{Peg\\Transfer}}
        & Pixel error                  & 323.6 $\pm$ 21.7 & 267.3 $\pm$ 19.7 & 241.9 $\pm$ 13.7 \\
        & Time in $\Omega_c$           & 43.0 $\pm$ 6.1   & 73.0 $\pm$ 7.2   & 71.4 $\pm$ 4.8   \\
        & Exits from $\Omega_{\text{out}}$ & 12.3 $\pm$ 2.4   & 14.1 $\pm$ 3.6   & 7.1 $\pm$ 2.4    \\
        & Time                         & 170.9 $\pm$ 24.8 & 150.1 $\pm$ 22.4 & 138.1 $\pm$ 14.8 \\
        \midrule
        \multirow{4}{*}{Suturing}
        & Pixel error                  & 202.6 $\pm$ 24.5 & 161.0 $\pm$ 20.9 & 155.1 $\pm$ 11.7 \\
        & Time in $\Omega_c$           & 59.9 $\pm$ 12.6  & 78.1 $\pm$ 17.4  & 79.2 $\pm$ 5.6   \\
        & Exits from $\Omega_{\text{out}}$ & 14.8 $\pm$ 10.6  & 18.2 $\pm$ 9.5   & 7.6 $\pm$ 3.2    \\
        & Time                         & 141.8 $\pm$ 44.1 & 150.4 $\pm$ 21.8 & 110.0 $\pm$ 26.8 \\
        \bottomrule    \end{tabular}
\end{table}

\subsubsection{Results and Discussion}
\label{subsec:exp3_results}

Table~\ref{tab:exp3_metrics} summarizes the task-level results across the three conditions. Compared with fixed-FoV control, the proposed method significantly reduces pixel error, increases the time inside \(\Omega_c\), nearly halves the exits from \(\Omega_{\text{out}}\), and shortens completion time in both tasks (all \(p < 0.001\), paired \(t\)-test). Compared with tracking-based control, the proposed method achieves lower pixel error, substantially fewer exits from \(\Omega_{\text{out}}\) (\(p < 0.001\)), and shorter completion times. Notably, tracking-based control yields even more \(\Omega_{\text{out}}\) exits than fixed FoV (and slower completion in Suturing), as uncoupled tracking errors under occlusion induce erratic camera movements that disrupt manipulation.

Fig.~\ref{fig:exp3_results}(a1) and (b1) present the spatial heatmaps of active entity positions under fixed FoV and the proposed control in Peg Transfer and Suturing. Compared with the dispersed distributions under fixed FoV, the proposed control concentrates the target distribution tightly around the image center. Consistently, the representative temporal error profiles in Fig.~\ref{fig:exp3_results}(a2) and (b2) show that the proposed method suppresses both horizontal and vertical deviations, maintaining lower pixel error and fewer cumulative exits throughout the trials.

Fig.~\ref{fig:exp3_results}(c) compares the tracking-based and proposed control methods during similar needle-passing scenes in Suturing. Without relational constraints during handover occlusion, the needle tracker in the baseline drifts onto the departing non-active instrument (or stalls in place) and pulls the aggregated FoV target away, whereas our semantic-coupled estimation accurately maintains the active instrument--needle association. Fig.~\ref{fig:exp3_results}(d) further compares the cumulative camera path length and computational performance of both strategies, showing that the proposed method significantly reduces the path length from $203.6\text{ mm}$ to $142.0\text{ mm}$ in Peg Transfer and from $132.5\text{ mm}$ to $79.3\text{ mm}$ in Suturing (\(p<0.001\)). Although the increased computational load leads to a slightly higher latency ($75.2\text{ ms}$ vs.\ $64.8\text{ ms}$) and a lower update frequency ($39.4\text{ Hz}$ vs.\ $45.0\text{ Hz}$), the system still maintains favorable real-time responsiveness. Overall, this demonstrates a favorable trade-off between control stability and computational overhead.


\section{Conclusion and Discussion}

In this letter, we proposed a semantic-coupled perception framework for autonomous laparoscope control that integrates surgical phase recognition, object detection and tracking, and camera-motion compensation. The key idea is to let phase-dependent relationships between surgical entities constrain state estimation when visual observations are ambiguous or unavailable. Benchmark and phantom experiments validated the framework in surgical phase recognition, target tracking, and autonomous FoV control, demonstrating improved target state estimation and more efficient task execution compared with the evaluated baselines. These results support coupling task-level semantics with visual state estimation for robotic laparoscope guidance.

A current limitation is that in vivo performance under soft-tissue deformation and respiration remains to be validated. Future work will evaluate the framework in vivo under more complex occlusions and scene changes and extend it to other surgical tasks with task-dependent entity relationships.






\bibliographystyle{IEEEtran}
\bibliography{references}

\end{document}